\documentclass[conf]{new-aiaa}
\usepackage[utf8]{inputenc}
\usepackage{subdepth}
\usepackage{graphicx}
\usepackage{verbatim}
\usepackage{amsmath}
\usepackage[version=4]{mhchem}
\usepackage{siunitx}
\usepackage{longtable,tabularx}
\usepackage{multicol}
\usepackage{booktabs}
\usepackage{subcaption} 
\usepackage{algorithm}
\usepackage{algpseudocode}
\usepackage{todonotes}
\usepackage{pythonhighlight}

\usepackage{tcolorbox}
\definecolor{ucsd_blue}{RGB}{24, 43, 73}
\definecolor{ucsd_gold}{RGB}{198, 146, 20}
\definecolor{ucsd_blue_light}{RGB}{0, 98, 155}
\definecolor{ucsd_gold_light}{RGB}{255, 205, 0}
\definecolor{ucsd_gray}{RGB}{116, 118, 120}

\title{Knowledge-guided generative surrogate modeling for high-dimensional design optimization under scarce data}

\author{Bingran Wang}
\affil{University of California San Diego, La Jolla, CA, USA}\affil{University of Sydney, Sydney, NSW, Australia}
\author{Seongha Jeong}
\affil{Samsung Electronics Co., Ltd, Republic of Korea}
\author{Sebastiaan P. C. van Schie}
\affil{University of California San Diego, La Jolla, CA, USA}
\author{Dongyeon Han, Jaeho Min}
\affil{Samsung Electronics Co., Ltd, Republic of Korea}
\author{John T. Hwang}
\affil{University of California San Diego, La Jolla, CA, USA}
\begin{document}
        
        




    

\maketitle
 \begin{abstract}
Surrogate models are widely used in mechanical design and manufacturing process optimization, where high-fidelity computational models may be unavailable or prohibitively expensive. 
Their effectiveness, however, is often limited by data scarcity, as purely data-driven surrogates struggle to achieve high predictive accuracy in such situations. Subject matter experts (SMEs) frequently possess valuable domain knowledge about functional relationships, yet few surrogate modeling techniques can systematically integrate this information with limited data.
We address this challenge with RBF-Gen, a knowledge-guided surrogate modeling framework that combines scarce data with domain knowledge. 
This method constructs a radial basis function (RBF) space with more centers than training samples and leverages the null space via a generator network, inspired by the principle of maximum information preservation. The introduced latent variables provide a principled mechanism to encode structural relationships and distributional priors during training, thereby guiding the surrogate toward physically meaningful solutions.
Numerical studies demonstrate that RBF-Gen significantly outperforms standard RBF surrogates on 1D and 2D structural optimization problems in data-scarce settings, and achieves superior predictive accuracy on a real-world semiconductor manufacturing dataset. 
These results highlight the potential of combining limited experimental data with domain expertise to enable accurate and practical surrogate modeling in mechanical and process design problems.
\end{abstract}
\section{Introduction}
In many mechanical design and manufacturing process optimization problems, high-fidelity physics-based simulation models are either unavailable or prohibitively expensive due to the complexity and proprietary nature of the underlying processes. 
As a result, engineers must often rely on experimental measurements or a limited number of simulations to make informed decisions on process designs. 
Such data are typically scarce, which makes the optimization problem particularly challenging--especially in high-dimensional settings.

To overcome these challenges, surrogate models are widely employed to approximate system responses from data. Their effectiveness in design optimization, however, depends critically on achieving high predictive accuracy despite data scarcity. Common surrogate modeling techniques include radial basis functions (RBF)~\cite{buhmann2000radial}, kriging~\cite{krige1951statistical, matheron1963principles}, inverse distance weighting (IDW)~\cite{shepard1968two}, and neural networks~\cite{rumelhart1986learning}. Each method provides unique advantages depending on the complexity, dimensionality, and noise characteristics of the problem.
These techniques have demonstrated success in a wide range of applications, including semiconductor manufacturing~\cite{wang2024design, zhou2011modeling}, aerospace system design ~\cite{cardoso2024constrained, han2012efficient, shen2025graph}, mechanical and materials design~\cite{gaul2009approximation, qi2021surrogate,zhang2020data, jiang2017data, lee2019machine,li2023deep}, and chemical process design~\cite{boukouvala2016predictive, singh2021machine}.
Across these domains, surrogate models enable engineers to explore complex design spaces, accelerate optimization, and reduce reliance on expensive experiments or simulations.

However, in practice, the performance of surrogate models often deteriorates when only a small amount of data is available. On one side, surrogate modeling methods such as radial basis functions (RBF) and kriging are known to perform well for low-dimensional problems with small datasets, but they lack mechanisms to directly incorporate engineering knowledge into the modeling process. 
In high-dimensional, data-scarce regimes, surrogates built solely from limited measurements often suffer from poor predictive accuracy, leading to suboptimal designs and minimal improvement in the quantities of interest. On the other side, neural network methods such as physics-informed neural networks (PINNs)~\cite{raissi2019physics} and generative adversarial networks (GANs)~\cite{goodfellow2014generative} can embed governing equations and distributional priors into training, thereby compensating for data scarcity. While these approaches reduce dependence on large datasets compared to purely data-driven neural networks, they remain limited in the types of prior knowledge they can incorporate and still face significant challenges in high-dimensional, data-scarce settings.

In many real-world settings, subject matter experts (SMEs) possess valuable domain knowledge about the functional relationships between design variables and process responses. This knowledge may include monotonicity with respect to certain variables, known output bounds, convexity or concavity, similarity to known analytical functions, or qualitative sensitivity trends. When properly incorporated into surrogate modeling, such information has the potential to substantially improve predictive accuracy and, in turn, the quality of optimization results. Moreover, these priors can guide the construction of a family of candidate functions that interpolate the available data, enabling more reliable predictions of design improvements. Despite this potential, no existing surrogate modeling framework systematically integrates such knowledge into the training process.

This paper addresses these challenges by introducing RBF--Gen, a novel surrogate modeling framework that integrates scarce data with domain knowledge. The method constructs an overcomplete radial basis function (RBF) representation and leverages a generative modeling approach to explore the null space of admissible interpolants, enabling the systematic incorporation of structural priors (e.g., monotonicity, convexity, and bounds) and distributional knowledge provided by domain experts. In contrast to existing physics-guided generative models, RBF--Gen does not rely on explicit governing equations or physical residuals. Instead, it embeds qualitative expert knowledge directly into the surrogate training process through penalty- and information-based objectives, ensuring exact interpolation of known data while allowing controlled variability consistent with physical intuition. By combining limited data with domain-informed priors, RBF--Gen serves as a natural extension of the standard RBF method, expanding its applicability to data-scarce problems where domain knowledge can be partially encoded as structural or distributional priors.

We evaluate the effectiveness of RBF-Gen on three case studies: (i) a one-dimensional cantilever beam optimization problem using simulation data, (ii) a two-dimensional shell thickness optimization problem using simulation data, and (iii) a semiconductor etching process surrogate modeling problem with real-world manufacturing data.
Across all cases, RBF-Gen consistently achieves higher predictive accuracy and enables larger design improvements compared to standard RBF surrogates in data-scarce settings.
These results demonstrate the value of combining limited experimental data with engineering knowledge, and highlight the potential of RBF-Gen as a practical and reliable surrogate modeling framework for complex mechanical design and manufacturing process problems.

The remainder of the paper is organized as follows. Section 2 reviews background on surrogate modeling, kernel methods, and generative adversarial networks. Section 3 presents the proposed RBF-Gen methodology. Section 4 reports numerical results on three case studies. Section 5 concludes with a summary and directions for future work.
\section{Background}
\label{sec: background}

\subsection{Surrogate modeling from scarce data}
We consider a single-output surrogate modeling problem with $N$ experimental measurements, denoted as
\[
\mathcal{D} = \{ (\mathbf{x}_i, y_i) \}_{i=1}^N,
\]
where $\mathbf{x}_i \in \mathbb{R}^d$ are the $d$-dimensional design or process variables, and $y_i \in \mathbb{R}$ is the corresponding quantity of interest (QoI).  
The objective is to construct a surrogate $\hat{f} : \mathbb{R}^d \to \mathbb{R}$ that accurately approximates the unknown mapping $f(\mathbf{x})$ based only on the limited dataset $\mathcal{D}$.  

In many mechanical and manufacturing applications, the input dimension $d$ can be large while the number of available data points $N$ is very small, owing to the high cost and long turnaround time of experiments.  
This imbalance between $d$ and $N$ makes surrogate construction severely ill-posed: the data admit many possible interpolants, yet only a few are physically meaningful or predictive outside the training region.
    
\subsection{Kernel-based surrogate modeling methods: RBF and kriging}
Kernel methods are widely used for surrogate modeling because they provide flexible, nonlinear approximations from scattered data.  
The core idea is to represent the response $f(\mathbf{x})$ in terms of pairwise similarities between input points, encoded by a kernel function $\phi(r)$ or $k(\mathbf{x}, \mathbf{x}')$.  
Two of the most popular kernel-based surrogates in engineering design optimization are radial basis function (RBF) interpolation and kriging (also known as Gaussian process regression).

\paragraph{Radial basis function (RBF)}
The RBF method approximates $f(\mathbf{x})$ as a linear combination of radially symmetric basis functions centered at the training points:
\begin{equation}
    \hat{f}(\mathbf{x}) = \sum_{i=1}^N w_i \, \phi\!\left( \lVert \mathbf{x} - \mathbf{x}_i \rVert \right),
    \label{eq:rbf-standard}
\end{equation}
where $w_i$ are interpolation weights and $\phi(r)$ is a radial kernel, such as the Gaussian
\begin{equation}
    \phi(r) = \exp(-\epsilon^2 r^2)
\end{equation}
or the thin-plate spline 
\begin{equation}
    \phi(r) = r^2 \log r.
\end{equation}
Enforcing $\hat{f}(\mathbf{x}_i) = y_i$
 for all training points leads to the square linear system
\begin{equation}
    \Phi w = y, 
    \quad \Phi_{ij} = \phi\!\left(\lVert \mathbf{x}_i - \mathbf{x}_j \rVert \right),
\end{equation}
which has a unique solution if $\Phi$ is non-singular.  

RBFs are effective in many settings but face two drawbacks in scarce-data regimes:  
(i) they yield a \emph{single} interpolant, preventing exploration of alternative functions equally consistent with the data, and  
(ii) the basis is tied to the data locations, which can limit generalization when the experimental points are sparse or unevenly distributed.

\paragraph{Kriging (Gaussian process regression)}
Kriging, or Gaussian process regression (GPR), models $f(\mathbf{x})$ as a Gaussian process:
\begin{equation}
    f(\mathbf{x}) \sim \mathcal{GP}\!\big(\mu(\mathbf{x}), \, k(\mathbf{x}, \mathbf{x}') \big),
\end{equation}
with mean function $\mu(\mathbf{x})$ and covariance kernel $k(\mathbf{x}, \mathbf{x}')$.  
The predictor at a new point $\mathbf{x}_*$ is
\begin{equation}
    \hat{f}(\mathbf{x}_*) = \mu(\mathbf{x}_*) 
    + k(\mathbf{x}_*, X) \, K^{-1} \, (y - \mu(X)),
\end{equation}
where $K_{ij} = k(\mathbf{x}_i, \mathbf{x}_j)$ encodes pairwise covariances.  
Kriging naturally provides uncertainty estimates and is widely used in design optimization for exploration–exploitation tradeoffs.  
However, its performance depends heavily on kernel hyperparameters and likelihood optimization, and in the small-$N$, high-$d$ regime it often suffers from overfitting or numerical instability.

\medskip
Together, RBF and kriging highlight both the promise and the limitations of kernel-based surrogates: while they interpolate sparse data well, they lack mechanisms to incorporate domain knowledge, and their predictive accuracy can degrade sharply when $N \ll d$.

\subsection{Information maximization in generative modeling}
The principle of \emph{information maximization} (InfoMax) originated in neural network research
as a way to learn informative internal representations by maximizing the mutual information between inputs and outputs~\cite{linsker1988self,bell1995information}.
This information–theoretic perspective led to the development of models such as independent component analysis and
the information bottleneck method~\cite{tishby2000information}, which emphasize learning representations that preserve
task–relevant information while filtering out noise.

Building on these ideas, Chen et al.~\cite{chen2016infogan} proposed the Information Maximizing Generative Adversarial Network (InfoGAN),
which encourages interpretable latent spaces by maximizing the mutual information between a subset of latent codes and the generated outputs:
\begin{equation}
I(c;\hat{f}_z) = H(c) - H(c\,|\,\hat{f}_z).
\end{equation}
This formulation allows controllable variations in generated samples and has inspired subsequent work on
mutual–information–based autoencoders and contrastive representation learning.

These methods have since found applications in areas such as computer vision~\cite{hjelm2019learning},  
natural language processing~\cite{oord2018representation}, mechanical design~\cite{chen2023gan} and aerodynamics~\cite{ruh2023airfoil}.
Despite these advances, InfoGAN and related neural-network–based models are not directly
suitable for surrogate modeling in engineering design problems,
particularly in data-scarce settings. 
First, stable training typically
requires thousands of samples, whereas only a few dozen experiments may be available in real-world mechanical design and manufacturing process problems. Second, they lack mechanisms to encode domain knowledge such
as monotonicity, positivity, or convexity, which frequently arise in
mechanical and manufacturing applications.

\section{Methodology}
\label{sec: methodology}

\begin{algorithm}[h!]
\caption{Training procedure of the RBF--Gen framework}
\label{alg:RBF-Gen}
\begin{algorithmic}[1]
\Require Training dataset $\mathcal{D} = \{(x_i, y_i)\}_{i=1}^{N}$, number of RBF centers $K > N$, kernel function $\phi(r)$, and expert-provided prior knowledge.
\vspace{4pt}
\Statex
\textbf{Step 1: Construct RBF basis and interpolation system.}
\vspace{4pt}
\State Place $K$ RBF centers $\{c_j\}_{j=1}^{K}$ uniformly or quasi-randomly in the design domain.
\State Form the interpolation matrix $\Phi_{ij} = \phi(\|x_i - c_j\|)$.
\vspace{4pt}
\Statex
\textbf{Step 2: Compute null space of the interpolation system.}
\vspace{4pt}
\State Solve for a particular solution $w_0$ such that $\Phi w_0 = y$.
\State Compute null-space basis $N \in \mathbb{R}^{K \times (K-N)}$ such that $\Phi N = 0$.
\State Represent admissible interpolants as $w = w_0 + N\alpha$, where $\alpha$ are free coefficients.
\vspace{4pt}
\Statex
\textbf{Step 3: Define generator for null-space exploration.}
\vspace{4pt}
\State Initialize a generator network $G(z; \theta)$ that maps latent variables $z \sim \mathcal{N}(0, I)$ to null-space coefficients $\alpha = G(z; \theta)$.
\State Each latent sample $z$ produces a valid interpolant $f_z(x) = \Phi(x)^{\top}(w_0 + N G(z))$.
\vspace{4pt}
\Statex
\textbf{Step 4: Incorporate prior knowledge.}
\vspace{4pt}
\State Define loss terms for structural priors and distributional priors.
\State Construct total loss following Equation~\ref{eqn:loss}.
\vspace{4pt}
\Statex
\textbf{Step 5: Train the generator.}
\vspace{4pt}
\State Train the generator neural network to optimize $\theta$.
\end{algorithmic}
\end{algorithm}

To overcome the limitations of existing surrogate modeling methods, we propose the RBF–Gen framework, which extends the classical radial basis function (RBF) formulation to incorporate prior domain knowledge. By combining data with domain knowledge from subject-matter experts, the framework enables the construction of more accurate and reliable surrogate models for design optimization, particularly in data-scarce and high-dimensional settings.
Constructing a surrogate model with the RBF-Gen framework consists of three steps:
\begin{enumerate}
\item Define RBF basis with $K$ basis functions
\item Compute the nullspace of the RBF basis after interpolating $N < K$ data points
\item Train nullspace coefficients with domain knowledge
\end{enumerate}
We cover these steps in the next sections.

\subsection{Relaxed RBF centers: building a richer function space}
Classical RBF interpolation ties each basis function center to a training point, yielding a unique interpolant. While this ensures exact data fit, it limits flexibility and prevents exploration of alternative functions equally consistent with the data. To overcome this limitation, we relax the formulation by decoupling the choice of RBF centers from the data locations.  

We place $K > N$ centers $\{c_j\}_{j=1}^K$ uniformly (or quasi-randomly) across the design domain and represent the surrogate as
\begin{equation}
    \hat{f}(\mathbf{x}) = \sum_{j=1}^K w_j \, \phi\!\left(\lVert \mathbf{x} - c_j \rVert \right),
    \label{eq:rbf-relaxed}
\end{equation}
where $\phi(r)$ is a radial kernel and $w_j$ are the RBF weights.  
Enforcing interpolation at the $N$ data points leads to the underdetermined system
\begin{equation}
\label{eq:rbf-underconstrained}
    \Phi w = y, 
\end{equation}
where
\begin{equation}
    \Phi_{ij} = \phi\!\left(\lVert x_i - c_j \rVert \right),
    \quad 
    \Phi \in \mathbb{R}^{N \times K},
\end{equation}
with $w \in \mathbb{R}^K$ the RBF weight vector and $y = [y_1, \dots, y_N]^\top$ the vector of data points located at $x = [x_i, \dots, x_N]$.

\subsection{RBF nullspace: avenue for integration of additional knowledge}
Since $K > N$, infinitely many solutions to equation~\eqref{eq:rbf-underconstrained} exist. Any solution can be expressed as
\begin{equation}
    w = w_0 + N \alpha,
\end{equation}
where $w_0$ is a minimum-norm particular solution to equation~\eqref{eq:rbf-underconstrained}, $N \in \mathbb{R}^{K \times (K-N)}$ is a basis for the null space of $\Phi$, and $\alpha \in \mathbb{R}^{K-N}$ is a free parameter vector.  
Each choice of $\alpha$ defines a distinct surrogate model $\hat{f}$ that interpolates the given data points.

This relaxed construction offers three key benefits:
\begin{itemize}
    \item \textbf{Flexibility:} The null space enables systematic generation of an entire family of admissible interpolants, rather than a single fixed function.
    \item \textbf{Improved generalization:} By distributing centers across the domain rather than restricting them to the data points, the surrogate covers unexplored regions more effectively.
    \item \textbf{Knowledge integration:} The free parameters $\alpha$ serve as controllable degrees of freedom, allowing prior domain knowledge to guide the selection of admissible interpolants.
\end{itemize}

\begin{figure*}[t]{\includegraphics[width= 1.0\linewidth]{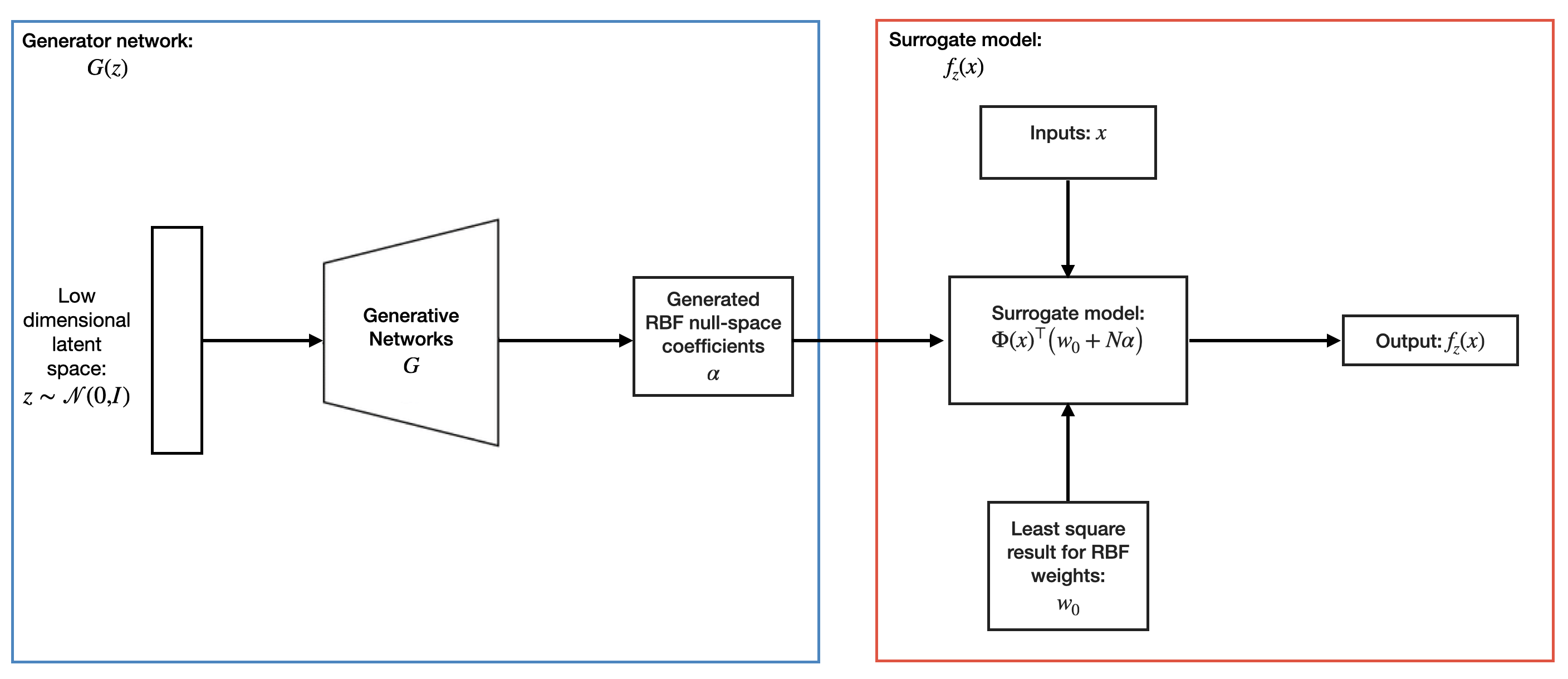}}
  \caption{Schematic illustration of the RBF-Gen framework.}
  \label{fig:rbf_gen}
\end{figure*}

\begin{figure*}[t]
  \centering
  \subfloat[\centering RBF with relaxed centers]{
    \includegraphics[width=.45\linewidth]{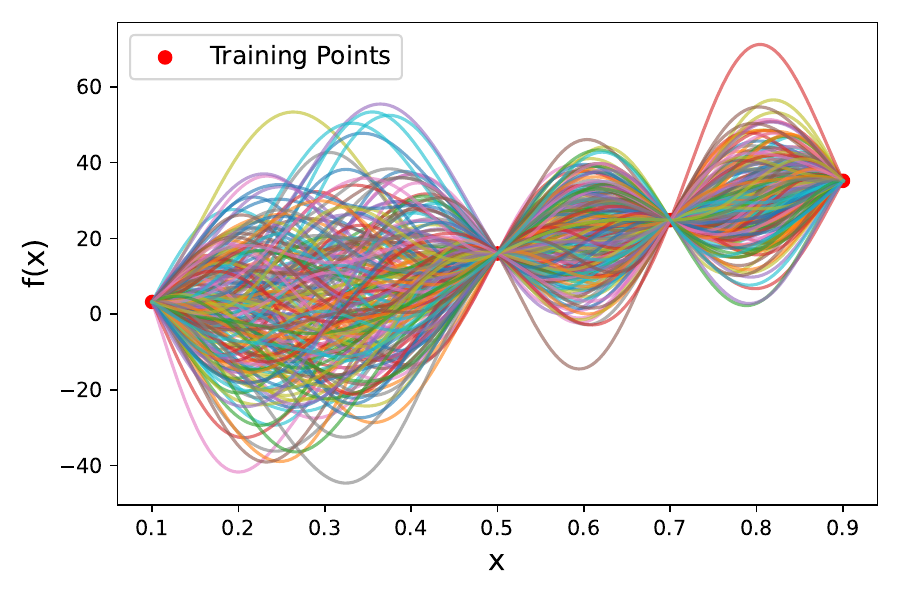}}
  \hfill
  \subfloat[\centering RBF-Gen with distribution prior: $f(0.3) \sim \mathcal{N}(\text{true}, 1)$]{
    \includegraphics[width=.45\linewidth]{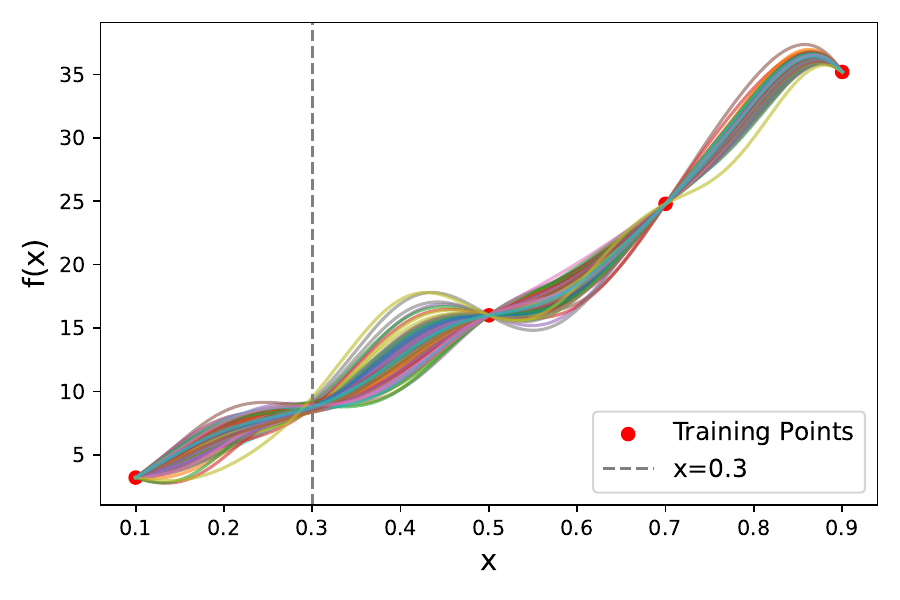}}
  \vskip\baselineskip
  \subfloat[\centering RBF-Gen with distribution prior: $\text{Hessian} \sim \mathcal{N}(\text{estimated}, 1)$]{
    \includegraphics[width=.45\linewidth]{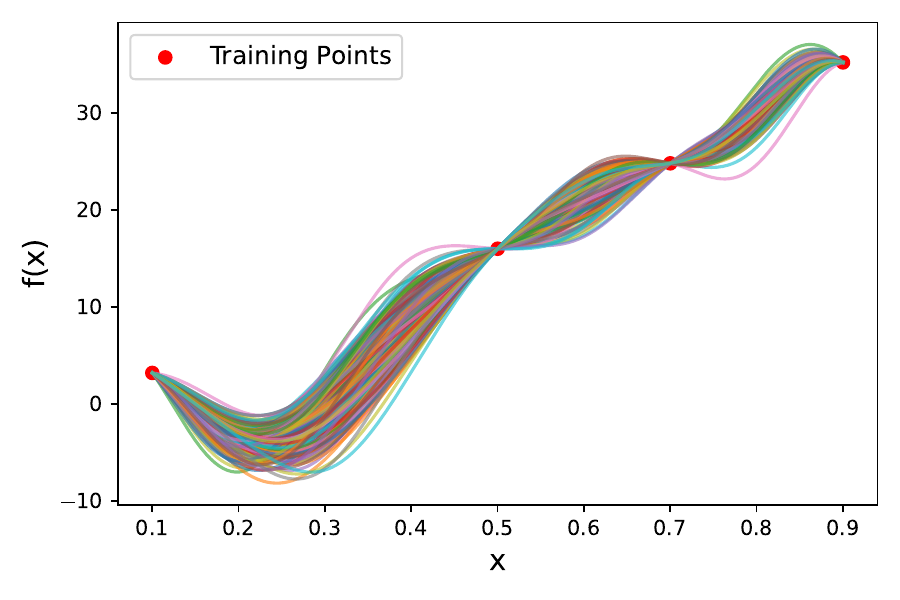}}
  \hfill
  \subfloat[\centering RBF-Gen with structural prior: monotonically increasing]{
    \includegraphics[width=.45\linewidth]{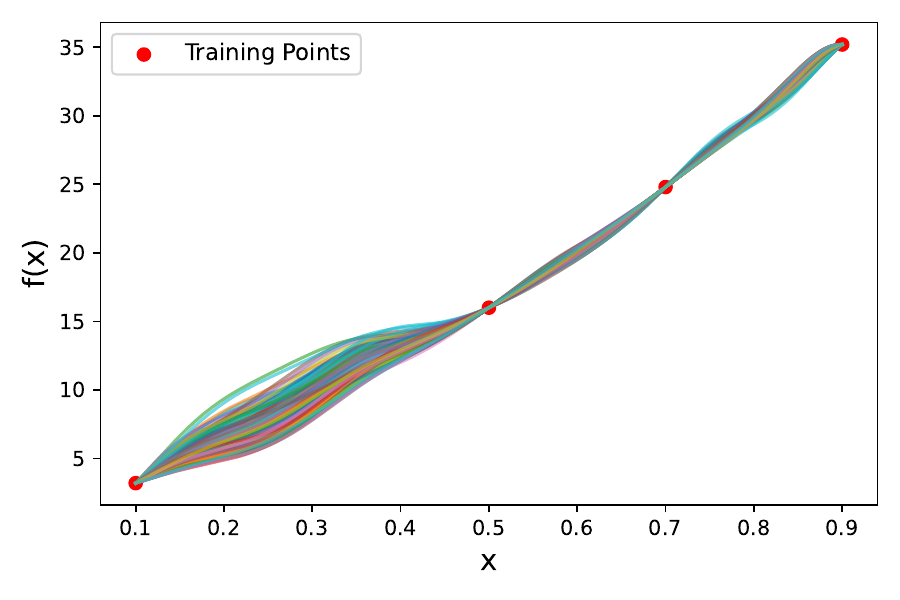}}
  \caption{RBF-generated functions with and without generator extensions.}
  \label{fig:infogan_results}
\end{figure*}

\subsection{RBF-Gen: Generator-based RBF surrogate modeling}

Building upon the RBF-based function generation framework, we extend it to incorporate prior knowledge using a generative modeling approach. Our goal is to generate functions that not only interpolate the known data points, but also exhibit desired statistical and structural characteristics that reflect domain-specific knowledge.

\subsubsection{Generator formulation}
We adopt a generator-based framework. 
The generator $G$ maps latent variables $z \sim \mathcal{N}(0,I)$ into coefficient vectors $\alpha$ that lie in the null space of the interpolation system:
\begin{equation}
\hat{f}_z(x) = \Phi(x)^\top \big( w_0 + N G(z) \big),
\end{equation}
where $w_0$ is a particular solution of the RBF interpolation system and $N$ is a basis for its null space.
Each latent sample $z$ thus yields a valid interpolant that exactly satisfies the training data, while the variability of $G(z)$ enables sampling an ensemble of admissible surrogates.

\subsubsection{Incorporating prior knowledge}
Unlike classical surrogate models, which rely solely on the training data, RBF-Gen is able to flexibly incorporate different sources of prior knowledge, from pointwise physical constraints to global distributional properties. Two complementary mechanisms are used to introduce different types of prior knowledge:  
(i) \emph{penalty terms}, which enforce structural properties such as monotonicity or positivity of $\hat{f}_z$, and  
(ii) \emph{divergence terms}, which align the distribution of generated function statistics with prescribed target distributions. 

The resulting penalty function for the generator is
\begin{equation}
\label{eqn:loss}
    \mathcal{L}_{\text{gen}} 
    = \sum_i \lambda_i \, \text{pen}_i(\hat{f}_z) 
    + \sum_j \gamma_j \, \mathrm{KL}\!\left(p_{\text{gen}}(s_j) \,\|\, p_{\text{target}}(s_j)\right),
\end{equation}
where $\text{pen}_i$ are soft structural penalty terms (e.g., violations of pointwise quantities such as monotonicity or positivity), $\mathrm{KL}$ denotes the Kullback--Leibler divergence, $s_j$ are functional statistics of interest (e.g., extrema, averages, curvature), and $\lambda_i,\gamma_j$ are tunable hyperparameters that allow one to place different weights on the various prior knowledge terms.

The resulting surrogate modeling approach, which we term \textbf{RBF-Gen}, produces ensembles of interpolants that both satisfy sparse data constraints and reflect expert-specified physical knowledge, without requiring adversarial discriminator training. A schematic illustration of the RBF--Gen framework is shown in Figure~\ref{fig:rbf_gen}, 
and the corresponding pseudocode outlining its main steps is provided in Algorithm~\ref{alg:RBF-Gen}.

\subsection{Different forms of information in RBF--Gen}
We give an overview of how various common types of domain knowledge can be formulated for use in the RBF--Gen framework:
\begin{itemize}
  \item \textbf{Penalty terms (structural priors)}  
  These act pointwise on generated functions $\hat{f}_z(x)$ and are implemented as soft penalties on violations:
  \begin{itemize}
    \item \emph{Monotonicity (increasing or decreasing):}
    \begin{equation}
    \text{pen}_{\text{mono}} = \frac{1}{|\mathcal{G}|}\sum_{x_k\in\mathcal{G}} 
    \text{ReLU}\!\big( \pm ( \hat{f}_z(x_{k+1}) - \hat{f}_z(x_k) ) \big)
    \label{eq:pen_mono}
    \end{equation}
    where $+$ enforces non-decreasing and $-$ enforces non-increasing trends.

    \item \emph{Positivity / bounds:}
    \begin{equation}
    \text{pen}_{\text{pos}} = \text{ReLU}\!\big(m - \min_{x\in\mathcal{G}} \hat{f}_z(x)\big)
    \label{eq:pen_pos}
    \end{equation}
    ensures $\hat{f}_z(x)\ge m$ across probe points $\mathcal{G}$.

    \item \emph{Lipschitz / slope bound:}
    \begin{equation}
    \text{pen}_{\text{Lip}} = \sum_{(x,y)\in\mathcal{P}}
    \text{ReLU}\!\left( \frac{|\hat{f}_z(x)-\hat{f}_z(y)|}{\|x-y\|} - L \right)
    \label{eq:pen_Lip}
    \end{equation}

    \item \emph{Smoothness / curvature:}
    \begin{equation}
    \text{pen}_{\text{curv}} = \sum_{x_k\in\mathcal{G}}
    \left( \hat{f}_z(x_{k+1}) - 2\hat{f}_z(x_k) + \hat{f}_z(x_{k-1}) \right)^2
    \label{eq:pen_curv}
    \end{equation}

    \item \emph{Convexity / concavity:}
    \begin{equation}
    \text{pen}_{\text{conv}} = \sum_{x_k\in\mathcal{G}}
    \text{ReLU}\!\big(-\Delta^2 \hat{f}_z(x_k)\big)
    \label{eq:pen_conv}
    \end{equation}
    enforces convexity ($\Delta^2 f \ge 0$); flip sign for concavity.

    \item \emph{Boundary / symmetry conditions:}
    \begin{equation}
    \text{pen}_{\text{bnd}} = \sum_{x_b\in\mathcal{B}} \big(\hat{f}_z(x_b) - v_b\big)^2
    \label{eq:pen_bnd}
    \end{equation}
    matches known boundary values $v_b$.
  \end{itemize}

  \item \textbf{KL divergence terms (distributional priors)}  
  These shape the distribution of function statistics across generated surrogates:
  \begin{itemize}
    \item \emph{Marginal values at a point $x_0$:}
    \begin{equation}
    \mathrm{KL}\!\big(p_{\text{gen}}(\hat{f}_z(x_0)) \,\|\, \mathcal{N}(\mu,\sigma^2)\big)
    \label{eq:KL_point}
    \end{equation}

    \item \emph{Regional averages:}
    \begin{equation}
    \mathrm{KL}\!\Big(p_{\text{gen}}\!\big(\tfrac{1}{|\mathcal{R}|}\sum_{x\in\mathcal{R}} \hat{f}_z(x)\big)\,\|\,p_{\text{target}}\Big)
    \label{eq:KL_region}
    \end{equation}

    \item \emph{Extremal values (max/min):}
    \begin{equation}
    \mathrm{KL}\!\big(p_{\text{gen}}(\max_{x\in\mathcal{R}} \hat{f}_z(x)) \,\|\, p_{\text{target}}\big)
    \label{eq:KL_extreme}
    \end{equation}

    \item \emph{Gradient magnitude distribution:}
    \begin{equation}
    \mathrm{KL}\!\big(p_{\text{gen}}(\|\nabla \hat{f}_z(x)\|) \,\|\, p_{\text{target}}\big)
    \label{eq:KL_grad}
    \end{equation}

    \item \emph{Curvature distribution:}
    \begin{equation}
    \mathrm{KL}\!\big(p_{\text{gen}}(\tfrac{\partial^2 \hat{f}_z}{\partial x_i^2}) \,\|\, p_{\text{target}}\big)
    \label{eq:KL_curv}
    \end{equation}

    \item \emph{Functional measures (integrals):}
    \begin{equation}
    \mathrm{KL}\!\big(p_{\text{gen}}(\int_\Omega g(x,\hat{f}_z,\nabla \hat{f}_z)\,dx) \,\|\, p_{\text{target}}\big)
    \label{eq:KL_integral}
    \end{equation}
  \end{itemize}
\end{itemize}

Through this design, RBF-Gen provides a unifying framework for combining sparse data with rich, heterogeneous domain knowledge, ensuring that generated interpolants remain both data-consistent and knowledge-consistent.

\subsection{Hyperparameter selection and tuning}

The performance and stability of the RBF--Gen framework depend critically on the appropriate choice of several hyperparameters, most notably the number of RBF centers ($K$) and the scaling factor for the null-space coefficients ($\alpha$). 
These parameters jointly control the expressiveness of the generated functions and the ability of the model to explore the admissible function space without overfitting the limited training data.

The parameter $K$ determines the size of the overcomplete RBF basis and directly influences the flexibility of the model. 
A larger $K$ increases the dimensionality of the null space, thereby enriching the diversity of generated functions; however, it can also make generator training more challenging due to higher computational cost and potential instability. 
In this work, we adopt a simple heuristic to select $K$ between two and five times the number of design variables ($D$),
which provides a practical balance between model expressiveness and numerical stability across the test cases studied.

The scaling factor for $\alpha$ controls the magnitude of the generator outputs projected into the null space of the RBF functions. 
This parameter determines the range of admissible function perturbations and, consequently, the diversity of generated surrogates. 
In practice, the scaling factor is selected to ensure sufficient variability in the generated ensemble while maintaining smoothness and physical consistency in the resulting function shapes.

\subsection{1D demonstration}

We demonstrate the effectiveness of the proposed RBF--Gen framework on a 1D function interpolation problem with scarce data.  
The underlying function is quadratic,  
\begin{equation*}
    f(x) = 20x^2 + 20x + 1,
\end{equation*}
but it is sampled at only 4 locations, making interpolation highly uncertain.  
Figure~\ref{fig:infogan_results} illustrates the outcomes of relaxed RBF interpolation and RBF--Gen when different forms of prior knowledge are incorporated into training.  
In particular, we consider priors based on target function values, curvature distributions, and monotonicity trends.  

The results show that relaxing the RBF centers yields a wide variety of interpolants that exactly match the training points, reflecting the uncertainty inherent in sparse data.  
With RBF--Gen, however, each type of prior shapes the generated functions to better reflect the expected structural or statistical properties of the true function.  
These findings highlight that systematically embedding prior knowledge enables RBF--Gen to produce interpolants that not only satisfy the data constraints but also align more closely with the underlying ground truth.
\section{Numerical Results}
\label{sec:results}
We demonstrate the effectiveness of the proposed RBF-Gen framework through three representative case studies of increasing complexity. 
First, we consider a \textbf{1D cantilever beam design optimization} problem, which serves as a simple yet illustrative benchmark for analyzing interpolation accuracy and design improvement under scarce data. 
Second, we study a \textbf{2D shell thickness optimization} problem, which introduces higher dimensionality and structural interactions, allowing us to assess the scalability of RBF-Gen to more complicated design problems. 
Finally, we apply RBF-Gen to a \textbf{semiconductor etching process} surrogate modeling problem, where only limited real-world experimental data is available. This case demonstrates the practicality of the proposed method in real-world manufacturing applications.
\subsection{1D cantilever beam design optimization}

\begin{figure*}[t]
    \centering
    \begin{subfigure}{0.48\linewidth}
        \centering
        \includegraphics[width=\linewidth]{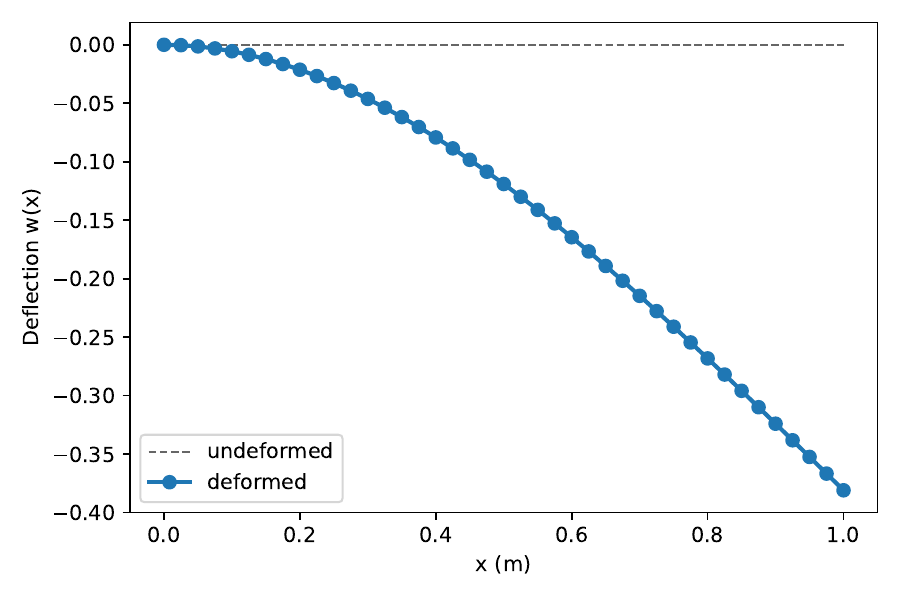}
        \caption{Deflected shape of the beam.}
        \label{fig:beam_deflection}
    \end{subfigure}
    \hfill
    \begin{subfigure}{0.48\linewidth}
        \centering
        \includegraphics[width=\linewidth]{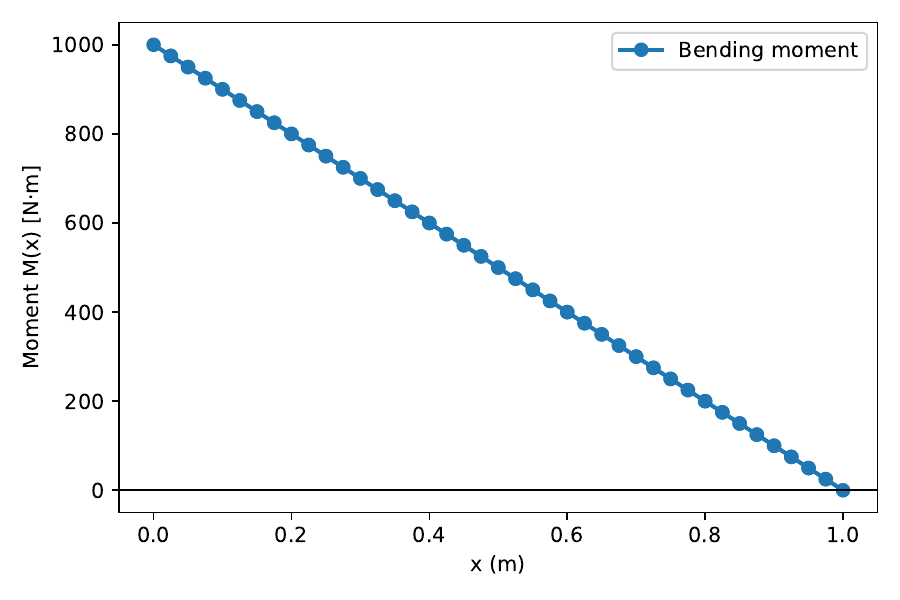}
        \caption{Bending moment distribution.}
        \label{fig:beam_moment}
    \end{subfigure}
    \caption{Structural response of a uniform cantilever beam.}
    \label{fig:beam_response}
\end{figure*}

\begin{figure*}[t]
    \centering
    \begin{subfigure}[t]{0.48\textwidth}
        \centering
        \includegraphics[width=\linewidth]{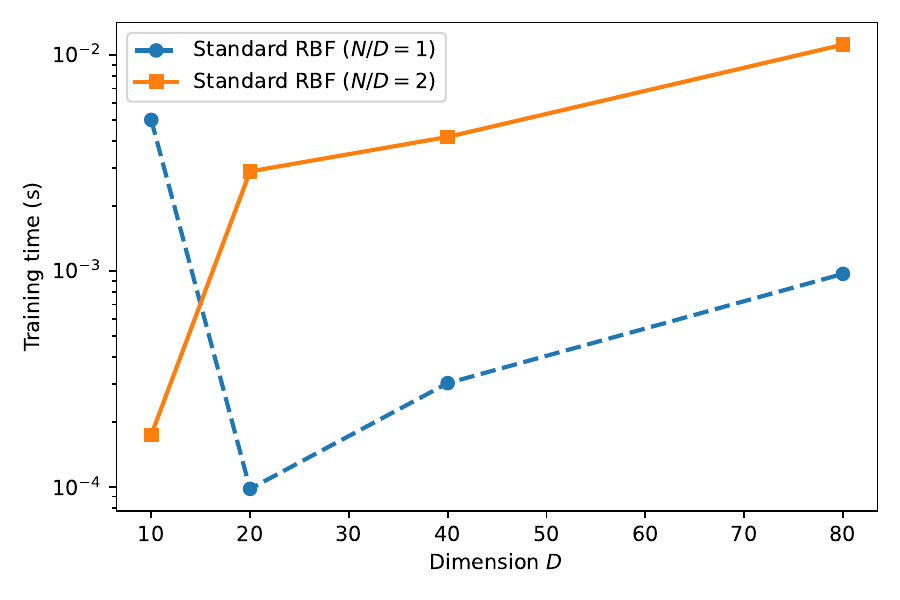}
        \caption{Standard RBF model.}
        \label{fig:rbf_training_time}
    \end{subfigure}
    \hfill
    \begin{subfigure}[t]{0.48\textwidth}
        \centering
        \includegraphics[width=\linewidth]{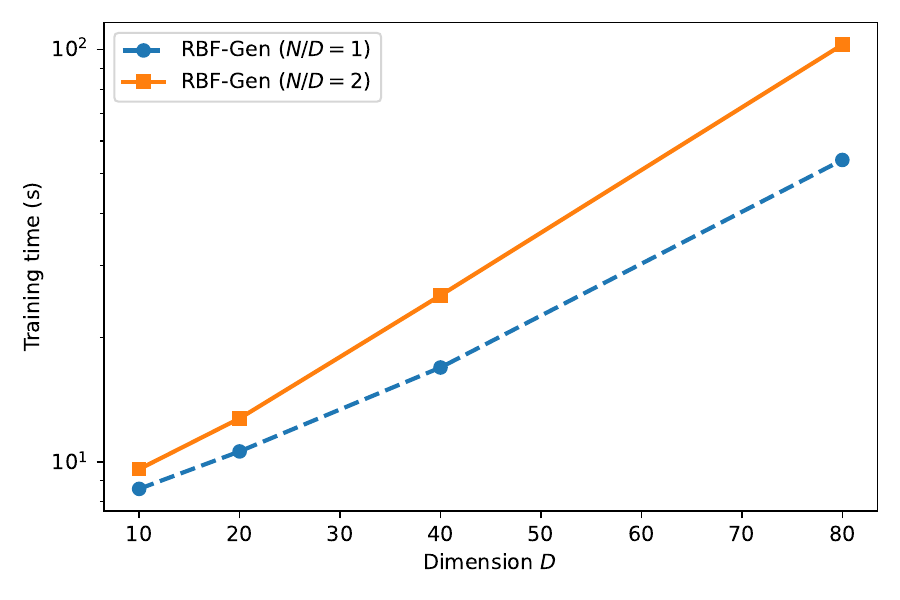}
        \caption{Proposed RBF--Gen model.}
        \label{fig:rbfgen_training_time}
    \end{subfigure}
    \caption{Comparison of training time versus problem dimension $D$ for (a) the standard RBF model and (b) the proposed RBF--Gen model under two data ratios ($N/D = 1$ and $N/D = 2$).}
    \label{fig:training_time_comparison}
\end{figure*}

\begin{figure*}[t]
    \centering
    \begin{subfigure}[t]{0.48\textwidth}
        \centering
        \includegraphics[width=\linewidth]{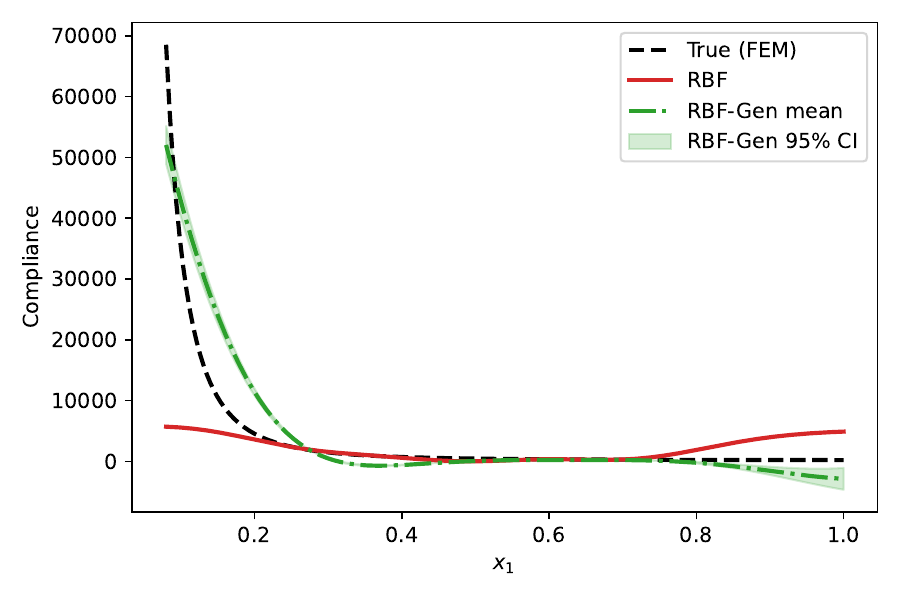}
        \caption{Slice plot for $D=2$, $N=4$.}
        \label{fig:cantilever_slice_d5n10}
    \end{subfigure}
    \hfill
    \begin{subfigure}[t]{0.48\textwidth}
        \centering
        \includegraphics[width=\linewidth]{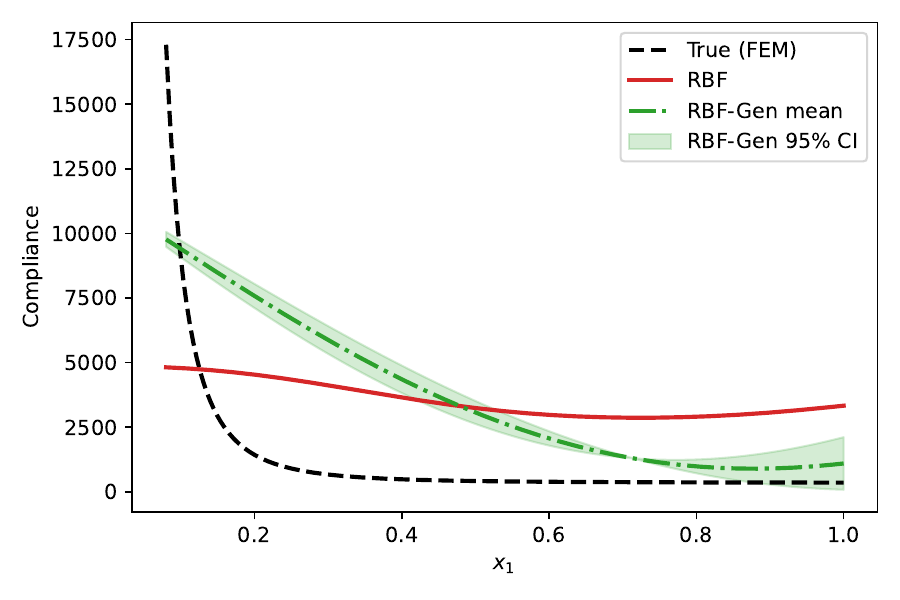}
        \caption{Slice plot for $D=10$, $N=20$.}
        \label{fig:cantilever_slice_d25n50}
    \end{subfigure}
    \caption{Slice plots of the cantilever compliance along $x_1$ comparing the true FEM response, standard RBF surrogate, and RBF--Gen surrogate (with 95\% CI).}

    \label{fig:cantilever_slice_comparison}
\end{figure*}

The first problem is a  structural design optimization problem, which minimizes the compliance of a cantilever beam under a point load, subject to a volume constraint. This problem provides a flexible benchmark with a known physics-based model, allowing us to evaluate the framework's ability to improve surrogate-driven optimization under data scarcity.

\subsubsection{Problem description}
In this problem, we have a cantilever beam that is discretized into $D$ finite elements along its length, and modeled using Euler--Bernoulli beam elements. Each element $i$ has a rectangular cross-section with fixed width $b$ and adjustable height $h_i$. The vector of element heights $\mathbf{h} = [h_1, \dots, h_D]^\top$ defines the design variables. The objective is to minimize the beam's tip compliance:
\begin{equation}
    C(\mathbf{h}) = \mathbf{f}^\top \mathbf{d},
\end{equation}
where $\mathbf{f}$ is the nodal load vector and $\mathbf{d}$ is the displacement vector obtained by solving the finite element equilibrium equations:
\begin{equation}
    \mathbf{K}(\mathbf{h}) \, \mathbf{d} = \mathbf{f}.
\end{equation}
The optimization problem contains a volume constraint,
\begin{equation}
    V(\mathbf{h}) = b \cdot \frac{L}{N} \sum_{i=1}^N h_i \leq 0.2,
\end{equation}
together with bound constraints on each element height,
\begin{equation}
    0.05 \leq h_i \leq 0.5.
\end{equation}

Figure~\ref{fig:beam_response} illustrates the structural response of a uniform cantilever beam in this problem.

\subsubsection{RBF-Gen implementation details}
To emulate realistic data scarcity, training samples of compliance values were generated from a small neighborhood of the design space, $0.05 \leq h_i \leq 0.1$, rather than spanning the entire feasible region. This reflects the practical setting where experimental data are often concentrated around nominal recipes. 

Regarding the domain knowledge, we incorporated three sources of information when training RBF-Gen:
\begin{itemize}
    \item \textbf{Monotonicity:} Compliance is expected to decrease as element thickness increases. For each design variable $x_j$, we constructed paired samples along its range while holding other variables fixed. Violations of the non-increasing trend were penalized.
    \item \textbf{Positivity:} Compliance values must remain strictly non-negative. A positivity penalty was introduced to ensure that all generated surrogates remain above zero across randomly probed points in the design domain.
    \item \textbf{Slice-based distributional priors:} For each design dimension, we selected one-dimensional slices of the design space (with other variables fixed at nominal values). At these slice points, compliance values were perturbed by $\pm 30\%$ around the finite-element "true" values to form target Gaussian distributions. The generated surrogate functions were then trained to match these target distributions by minimizing a KL divergence term.
\end{itemize}

The complete loss function for training the generator is
\begin{align}
\label{eq:lossfunction_beamshell}
\mathcal{L} \;=\;& 
  \lambda_{\text{mono}}\;
  \underbrace{\sum_{j=1}^{N}\sum_{k=1}^{G-1}
  \max\!\bigl(0,\;\Delta f_{j}^{(k)}\bigr)}_{\mathcal{L}_\text{mono}} \nonumber \\[6pt]
&+\;\lambda_{\text{pos}}\;
  \underbrace{\sum_{x\in \mathcal{G}} \max\!\bigl(0,\; -f(x)\bigr)}_{\mathcal{L}_\text{pos}} \nonumber \\[6pt]
&+\;\lambda_{\text{KL}}\;
  \underbrace{
  \mathrm{KL}\!\left(
    p_{\text{gen}}(f(x)) \,\big\|\, 
    \mathcal{N}\!\left(\mu_{t},\,\sigma_{t}^{2}\right)
  \right)}_{\mathcal{L}_\text{dist}} ,
\end{align}
where $\mu_{t}, \sigma_{t}$ are the prescribed slice-based statistics derived from the true finite element evaluations, and $p_{\text{gen}}$ is the empirical distribution induced by the generator across the sampled slices.

\subsubsection{Results and discussion}

\begin{figure*}[t]
    \centering
    \begin{subfigure}{0.48\linewidth}
        \centering
        \includegraphics[width=\linewidth]{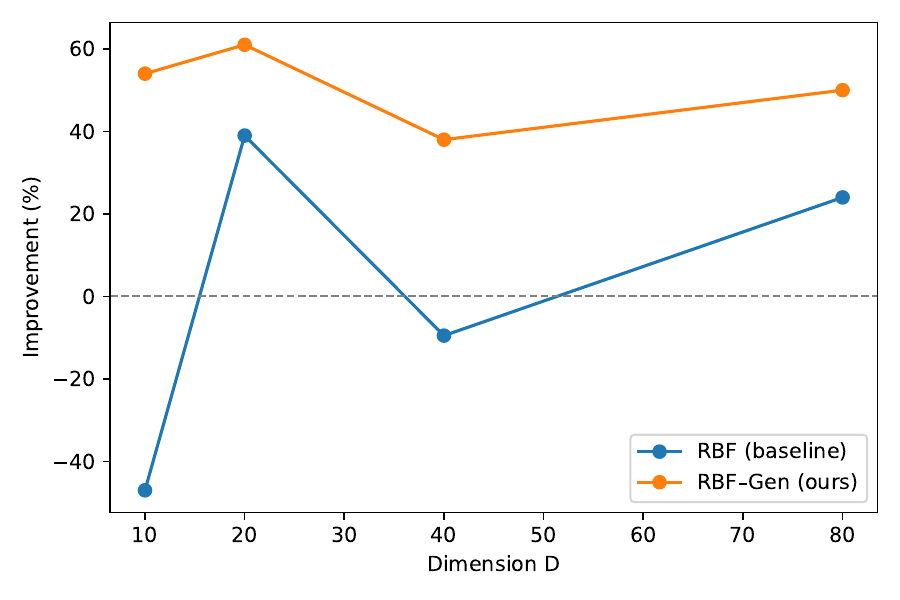}
        \caption{Case $N/D = 1$.}
        \label{fig:beam-improvement-nd1}
    \end{subfigure}
    \hfill
    \begin{subfigure}{0.48\linewidth}
        \centering
        \includegraphics[width=\linewidth]{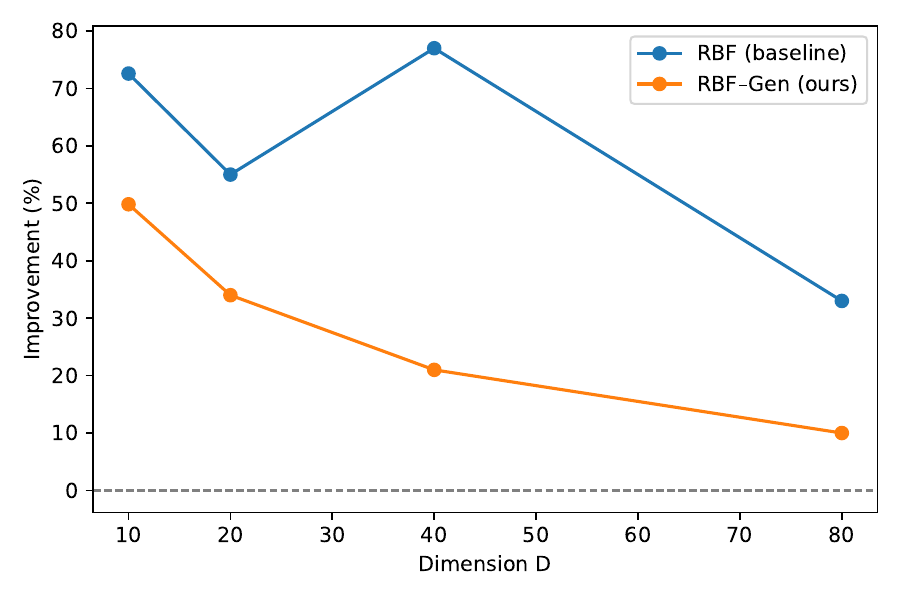}
        \caption{Case $N/D = 2$.}
        \label{fig:beam-improvement-nd2}
    \end{subfigure}
    \caption{Measured improvement (\%) in the 1D cantilever beam optimization problem as a function of design dimension $D$. 
    Here, $N$ denotes the number of training samples used to fit the surrogate models and $D$ denotes the number of design variables (beam elements). 
    Results are shown for (a) $N/D=1$ and (b) $N/D=2$.}
    \label{fig:beam-improvement}
\end{figure*}

The training time comparison between the standard RBF and the proposed RBF--Gen models is shown in Figure~\ref{fig:training_time_comparison}. As expected, the RBF--Gen model requires considerably higher computational time than the standard RBF model due to the additional steps of null-space computation and generator training. Moreover, the training time of RBF--Gen increases superlinearly with the problem dimension, reflecting the higher computational cost associated with exploring the admissible function space in higher dimensions. Nevertheless, in the context of practical mechanical and manufacturing process optimization problems, this additional computational overhead is typically negligible. When each experimental measurement or high-fidelity simulation can take hours or even days to complete, the training time of RBF--Gen, on the order of seconds to minutes, becomes trivial relative to the total cost of data generation.

The slice plots in Figure~\ref{fig:cantilever_slice_comparison} illustrate the predicted compliance profiles along the first design variable, comparing the true finite element (FEM) response, the standard RBF surrogate, and the proposed RBF--Gen model. Each slice represents the predicted response when varying a single variable while fixing the others at their mean values, thereby providing an interpretable view of how each surrogate captures the underlying functional dependence. The results show that the functions generated by the RBF--Gen model follow the true FEM response much more closely than those obtained from the standard RBF surrogate, demonstrating that the inclusion of domain knowledge leads to more physically consistent predictions. However, as the dimensionality increases from $D=2$ to $D=10$, the generated functions deviate further from the true response, reflecting the increasing difficulty of accurately modeling high-dimensional problems with limited data. Despite this degradation, the RBF--Gen model continues to outperform the standard RBF approach, maintaining better alignment with the true physical trend.

Regarding the effectiveness of the surrogate models in design optimization, we use the \textit{measured true improvement} as the  performance metric. The measured true improvement quantifies the actual reduction in the objective function achieved through optimization based on each surrogate model, evaluated against a common initial design point. Specifically, for each case, surrogate-based optimization is performed using either the standard RBF or the proposed RBF--Gen model, and the resulting optimal design is re-evaluated using the true finite element model. The percentage reduction in the true objective function relative to the initial design is then reported as the measured true improvement.
Figure~\ref{fig:beam-improvement} summarizes the measured true improvement (\%) across different design dimensions $D$ for two levels of data availability: 
(i) $N/D=1$, where the number of training samples ($N$) equals the number of design variables ($D$), representing a highly data-scarce regime, and 
(ii) $N/D=2$, where twice as many training samples as design variables are available.

In the scarce-data setting ($N/D=1$, Fig.~\ref{fig:beam-improvement-nd1}), the proposed RBF-Gen framework consistently outperforms the standard RBF surrogate. 
Across dimensions ranging from $D=10$ to $D=80$, RBF--Gen achieves substantially higher design improvements.
These results highlight the strength of RBF-Gen in leveraging prior knowledge to guide surrogate training when data are extremely limited.

In contrast, when more training samples are available ($N/D=2$, Fig.~\ref{fig:beam-improvement-nd2}), the performance of RBF--Gen deteriorates relative to the baseline RBF. 
Here, the baseline RBF achieves stronger improvements on average, while RBF--Gen tends to underperform. 
This behavior reflects the fact that, with richer data, in this problem, the standard RBF surrogate can already capture the input--output mapping effectively, while the additional priors imposed by RBF--Gen may introduce unnecessary biases.

Overall, these results demonstrate that, in this problem, RBF--Gen is particularly beneficial in data-scarce regimes, where classical surrogates suffer from poor generalization. 
However, as the data-to-dimension ratio improves, conventional RBF interpolation regains its advantage due to the direct availability of sufficient data to constrain the surrogate model.



\subsection{2D cantilever shell design optimization}

\begin{figure}[t]
    \centering
    \includegraphics[width=0.9\linewidth]{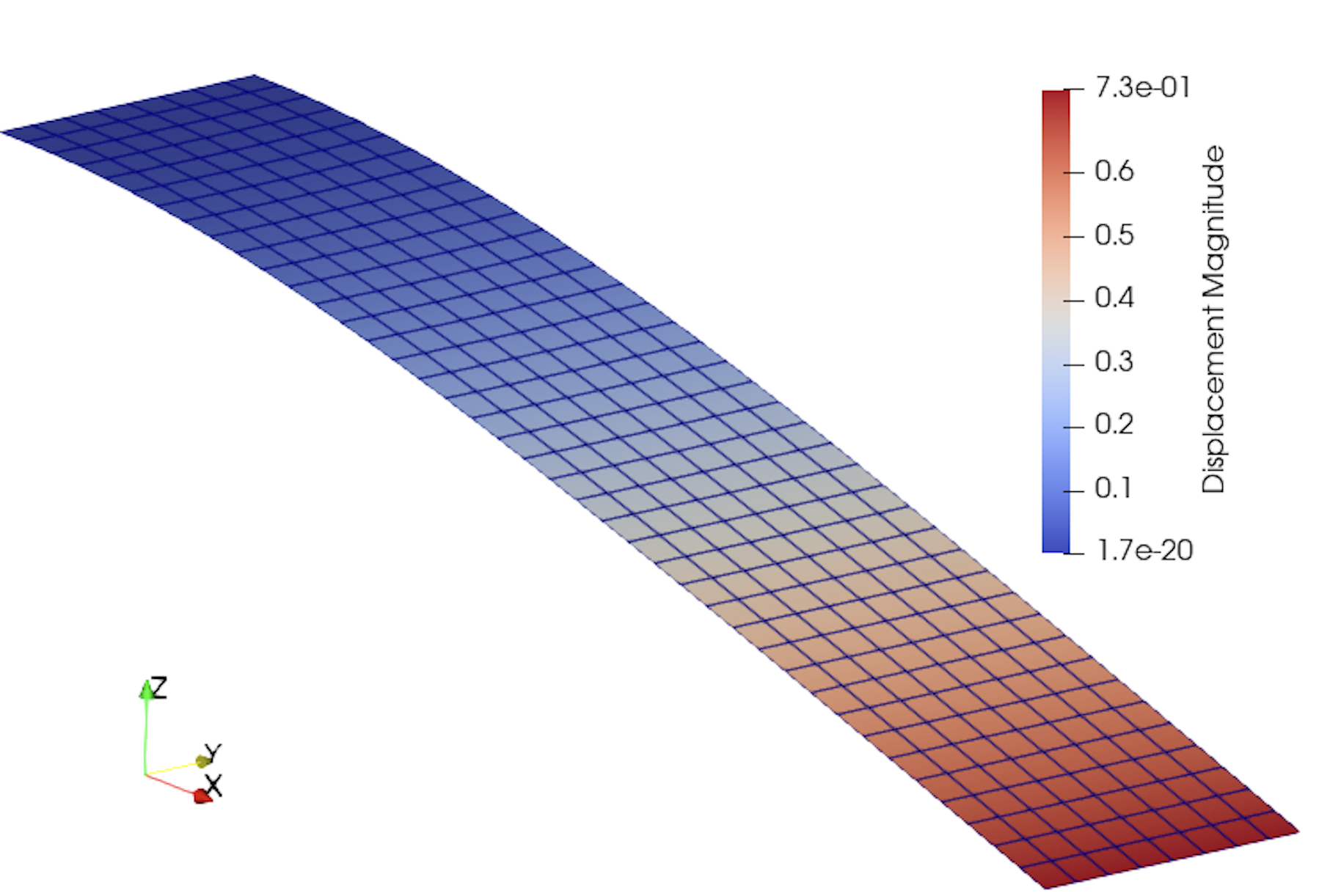}
    \caption{Structural response of a  cantilever thin plate subject to a distributed vertical load.}
    \label{fig:shell_deflected}
\end{figure}

The second problem is a shell thickness optimization problem. We model the steady-state deformation of a cantilever thin plate with the Reissner--Mindlin (RM) shell model. This test case allows us to demonstrate RBF-Gen's performance in optimization problems with more than one spatial dimension and high-dimensional optimization variable spaces.

\subsubsection{Problem description} 
We model the steady-state deformation of a $2\ m \ \times 10\ m$ cantilever plate using the Reissner--Mindlin (RM) shell model using FEniCS. Previously, FEniCS-based \cite{bib:Baratta2023} steady-state and time-dependent implementations of the RM shell model have been used for design optimization in \cite{bib:Xiang2024, bib:vanSchie2025a, bib:vanSchie2025b}. We refer the interested reader to \cite{bib:Campello2003} for a detailed description of the RM shell model.

For this problem, the plate is clamped on one edge and subjected to a distributed vertical load on the opposite edge. The shell thickness is taken as the design variable, discretized over the mesh elements of the domain $\Omega$ as a linearly-varying function.
The objective is to minimize the compliance,
\begin{equation}
    C(\mathbf{t}) = \mathbf{f}^\top \mathbf{d},
\end{equation}
where $\mathbf{f}$ is the applied nodal force vector representing a distributed vertical load, and $\mathbf{d}$ is the displacement vector obtained from the RM equilibrium equations. 
As in the beam case, the optimization problem includes a volume constraint,
\begin{equation}
    V(\mathbf{t}) = \sum_{i=1}^{N} \int_{\Omega} t_i \phi_i\ d\Omega \leq V_{\max},
\end{equation}
where $t_i \in \left[0.01, 0.2 \right]$ is the $i^{\text{th}}$ thickness degree of freedom and $\phi_i$ its corresponding finite element basis function. $V_{\max}$ is set equal to the volume of a plate with uniform thickness of $0.1\ m$, so $V_{\max} = 2$.

\subsubsection{RBF-Gen implementation details}
The RBF-Gen setup follows the same formulation as in the cantilever beam problem. 
Training data are sampled from a small region of the design space, and prior knowledge is incorporated through a positivity penalty on compliance and monotonicity priors on the thickness variables. 
These additions ensure the generated surrogates remain physically consistent while interpolating sparse training data.

\subsubsection{Results and discussion}
\begin{figure*}[t]
    \centering
    \begin{subfigure}{0.48\linewidth}
        \centering
        \includegraphics[width=\linewidth]{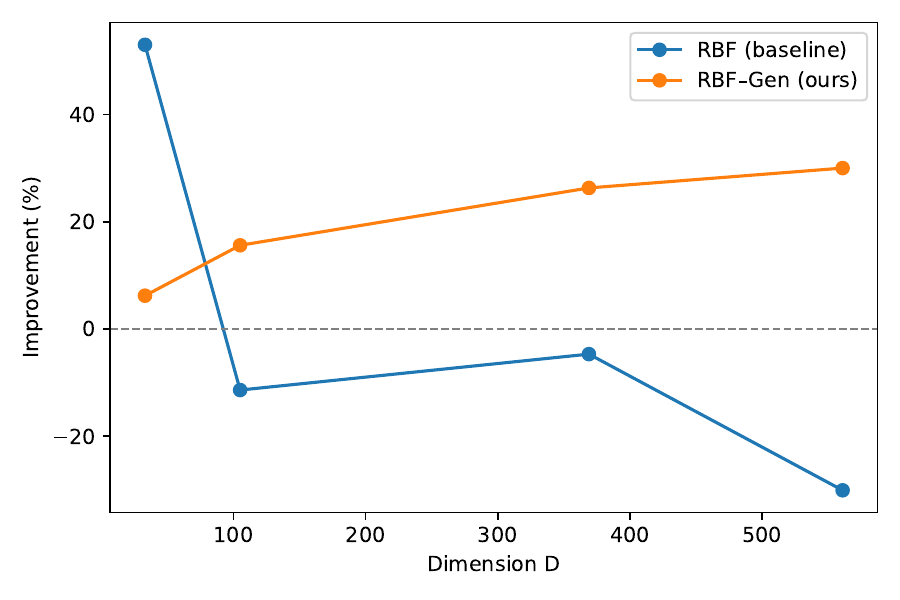}
        \caption{Case $N/D = 1$.}
        \label{fig:shell-improvement-nd1}
    \end{subfigure}
    \hfill
    \begin{subfigure}{0.48\linewidth}
        \centering
        \includegraphics[width=\linewidth]{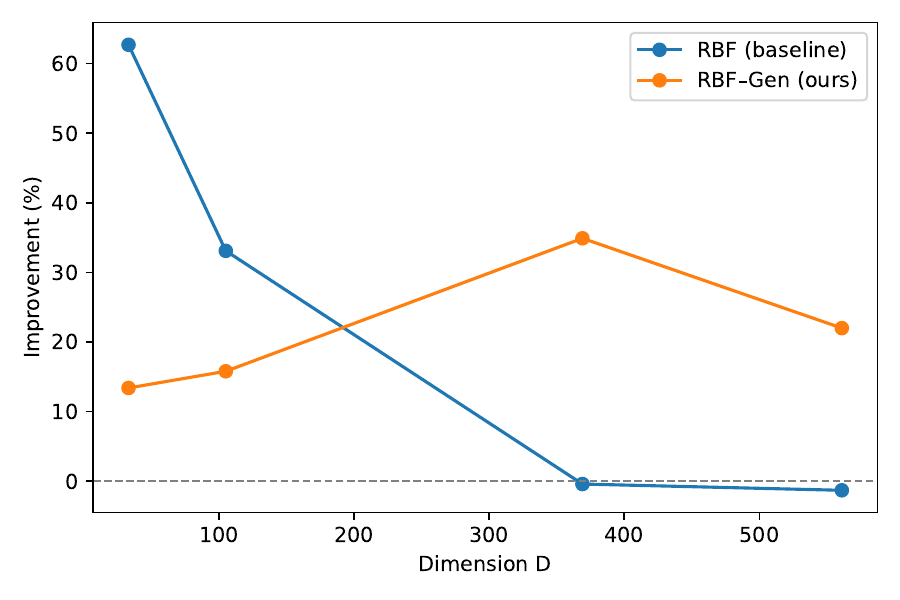}
        \caption{Case $N/D = 2$.}
        \label{fig:shell-improvement-nd2}
    \end{subfigure}
    \caption{Measured improvement (\%) in the 2D cantilever shell optimization problem as a function of design dimension $D$. 
    Here, $N$ denotes the number of training samples used to fit the surrogate models and $D$ denotes the number of design variables (shell elements). 
    Results are shown for (a) $N/D=1$ and (b) $N/D=2$.}
    \label{fig:shell-improvement}
\end{figure*}

We again compare the optimization performance of a standard RBF surrogate approach with the proposed RBF-Gen. We start by looking at the optimization improvements that are obtained when we have a single data point per design variable ($N/D=1$). Figure~\ref{fig:shell-improvement-nd1} shows that the RBF surrogate drastically degrades in performance as the number of design variables is increased: The RBF surrogate finds an optimal design that performs worse than the initial design for all but the smallest considered case. This in contrast to RBF-Gen, which outperforms the RBF surrogate for all but the lowest-dimensional optimization problem considered, and consistently finds optimized designs that outperform the initial design.

Figure~\ref{fig:shell-improvement-nd2} shows how both models perform when we double the number of available data points ($N/D=2$). While the performance of the RBF approach has improved relative to the results shown in Figure~\ref{fig:shell-improvement-nd1}, the RBF surrogate still fails to obtain any performance improvement in the higher-dimensional optimization cases. In contrast, RBF-Gen consistently finds optimal design points that improve upon the initial design. These results show that, in this problem using RBF-Gen has clear advantages for surrogate-based optimization over using a traditional RBF approach.

\begin{figure*}[t]{\includegraphics[width= 1\linewidth]{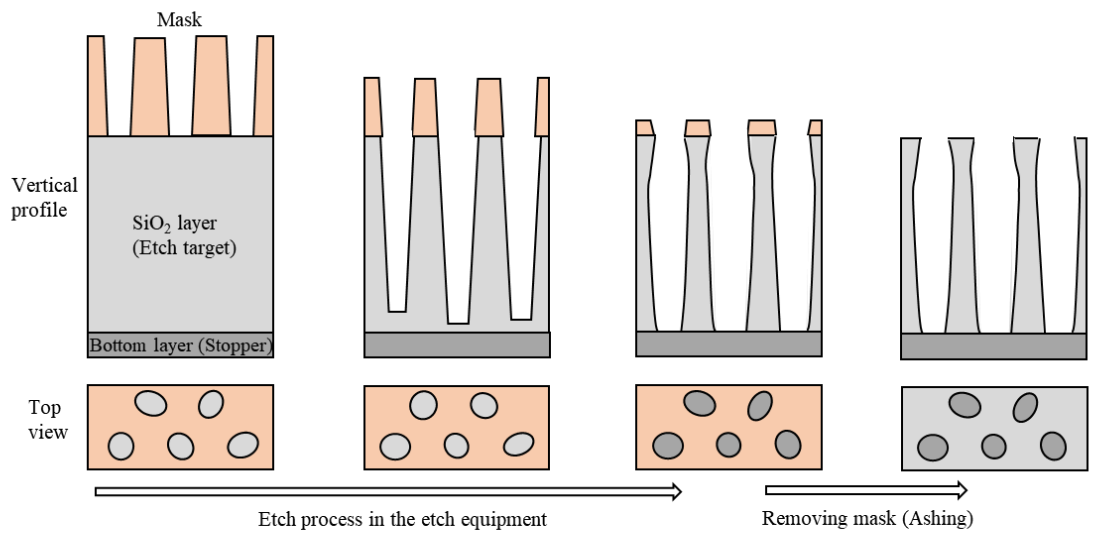}}
  \caption{ Vertical profiles and top views during the etch and ash processes}
  \label{fig:etch-process}
\end{figure*}

\begin{figure*}[t]{\includegraphics[width= 1\linewidth]{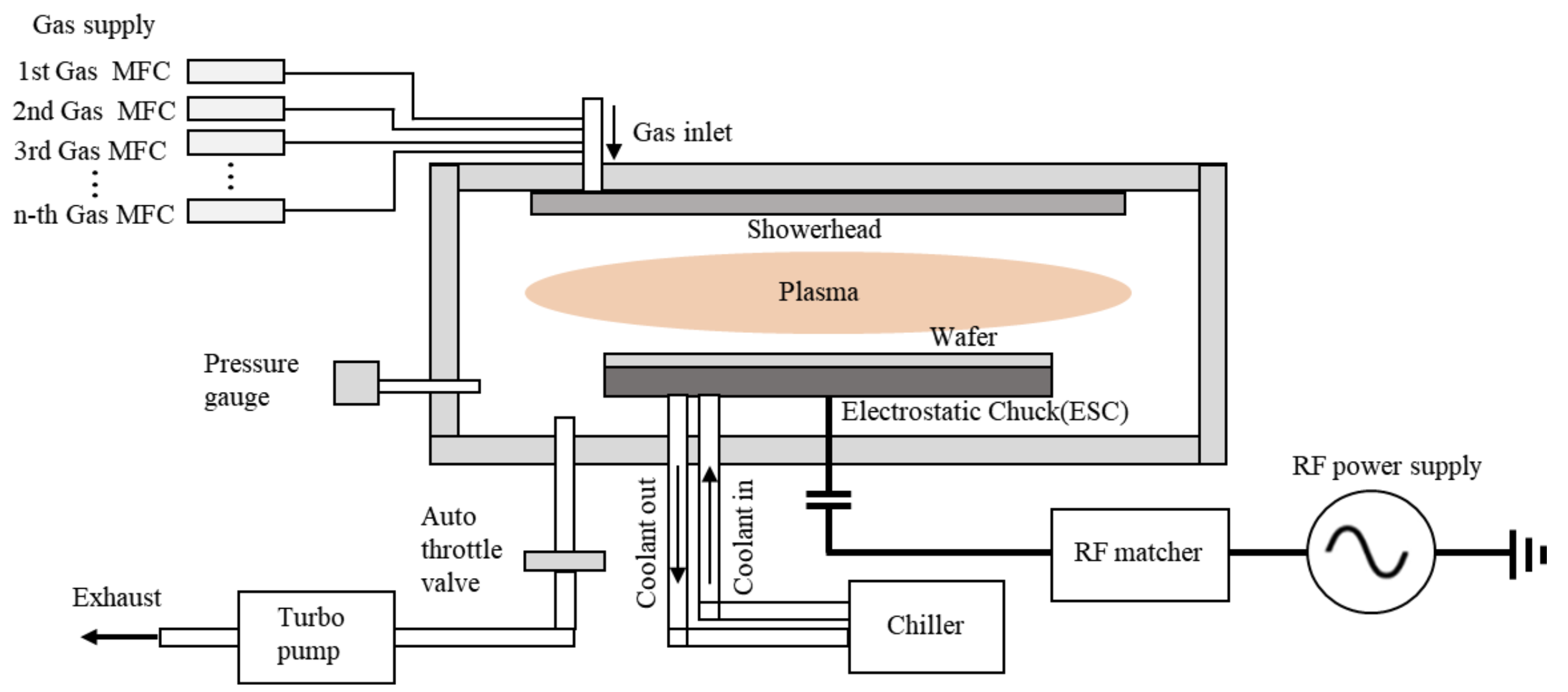}}
  \caption{Etching process equipment schematic}
  \label{fig:etch-photos}
\end{figure*}



\begin{table*}[t]
\centering
\caption{Monotonicity information provided by field experts. 
A “+1” indicates increasing, “–1” indicates decreasing, and “0” indicates no prior information. 
Input and output names are redacted for confidentiality.}
\label{tab:tab:mono-redacted}
\setlength{\tabcolsep}{15pt} 
\renewcommand{\arraystretch}{1.2}
\begin{tabular}{lccccc}
\toprule
\textbf{Variable} & \textbf{QoI 1} & \textbf{QoI 2} & \textbf{QoI 3} & \textbf{QoI 4} & \textbf{QoI 5} \\
\midrule
Variable 1  & +1 & -1 & -1 & +1 & -1 \\
Variable 2  &  0 &  0 &  0 &  0 &  0 \\
Variable 3  &  0 &  0 &  0 &  0 &  0 \\
Variable 4  & -1 & +1 & +1 & -1 & -1 \\
Variable 5  &  0 &  0 &  0 &  0 &  0 \\
Variable 6  & -1 & +1 & +1 & -1 & +1 \\
Variable 7  & -1 & +1 & +1 & -1 &  0 \\
Variable 8  & -1 & +1 & +1 & -1 &  0 \\
Variable 9  & -1 & +1 & +1 & -1 & +1 \\
Variable 10 &  0 & -1 & -1 & +1 &  0 \\
Variable 11 &  0 & -1 & -1 & +1 &  0 \\
Variable 12 &  0 & -1 & -1 & +1 & +1 \\
Variable 13 &  0 &  0 &  0 &  0 &  0 \\
Variable 14 & -1 & +1 & +1 & -1 & -1 \\
Variable 15 & -1 & +1 & +1 & -1 & -1 \\
Variable 16 &  0 &  0 &  0 &  0 &  0 \\
Variable 17 & -1 & +1 & +1 & -1 & -1 \\
\bottomrule
\end{tabular}
\end{table*}

\begin{table*}[t]
\centering
\caption{Leave-Two-Out (L2O) errors per QoI and overall means. 
Reported are average relative error (ARE) and average absolute error (AAE).}
\label{tab:l2o-results}
\setlength{\tabcolsep}{10pt} 
\renewcommand{\arraystretch}{1.2}
\begin{tabular}{lcccc}
\toprule
 & \multicolumn{2}{c}{\textbf{RBF (baseline)}} & \multicolumn{2}{c}{\textbf{RBF–InfoGAN (ours)}} \\
\cmidrule(lr){2-3}\cmidrule(lr){4-5}
\textbf{QoI} & ARE & AAE & ARE & AAE \\
\midrule
QoI~1 & 0.4140 & 6.9710  & 0.3220 & 5.4200 \\
QoI~2 & 0.1678 & 19.7025 & 0.3885 & 20.3100 \\
QoI~3 & 0.0940 & 7.5008  & 0.4505 & 8.9500 \\
QoI~4 & 1.4881 & 18.6276 & 0.6021 & 10.8200 \\
QoI~5 & 0.6544 & 5.8356  & 0.2500 & 4.1100 \\
\midrule
\textbf{Overall mean} & \textbf{0.5638} & \textbf{11.7275} & \textbf{0.5030} & \textbf{9.9220} \\
\bottomrule
\end{tabular}
\end{table*}

\subsection{Etching process data from semiconductor manufacturing }

We evaluate the proposed RBF--Gen on a real-world dataset from the development of a High Aspect Ratio Contact (HARC) etching process in semiconductor manufacturing. 
The HARC etch has become a critical step as the design rule (DR) in DRAM continues to shrink and the string height in 3D NAND devices increases~\cite{choe2021memory}, requiring deep and narrow holes to be etched with uniform diameters and vertical profiles.
Figure~\ref{fig:etch-process} illustrates a typical hole-etching process, in which a wafer coated with an SiO$_2$ layer and patterned photoresist (PR) mask is etched to produce high-aspect-ratio holes. 
Achieving the desired uniformity is extremely challenging due to the \emph{depth-loading effect}, where the etch rate decreases as the aspect ratio increases~\cite{yen2005profile}, and the mask-induced \emph{neck effect}, which causes a bowing region near the top of the hole profile~\cite{lee2010mechanism}. 
These coupled effects lead to significant shape distortion, where the upper region becomes wider and the lower region narrower, making it difficult to maintain a vertical profile throughout the depth. 
To compensate for these nonlinear and competing phenomena, process engineers must iteratively adjust multiple recipe parameters--including process time, power, temperature, pressure, and gas flow rate--across several sequential steps. 
However, each experimental trial is expensive and time-consuming, resulting in very few measured data points relative to the number of controllable variables. 
This imbalance creates a severely scarce-data setting in a high-dimensional input space, posing a major challenge for constructing reliable surrogate models capable of assisting in process recipe optimization.

The dataset used in this study comprises 34 real fabrication experiments, each corresponding to a complete multi-step etch recipe. 
Each recipe is defined by 17 controllable input variables that characterize the physical conditions inside the chamber. 
These inputs include the process times of each recipe step, the RF power and temperature setpoints, the chamber pressure, and the flow rates of multiple process gases. Together, these variables capture the knobs available to process engineers for manipulating plasma--material interactions during the etch.
Figure~\ref{fig:etch-photos} shows the schematic of the etching process equipment.

The process outputs are measured as 5 quantities of interest (QoIs), which summarize critical aspects of etch quality. 
The primary objective QoI is the average critical dimension (CD) difference, defined as the difference between the largest and smallest diameters along the vertical profile of an etched hole. Minimizing this difference ensures uniformity and reduces the risk of electrical malfunction in the fabricated device. 
The remaining four QoIs act as constraints that evaluate different aspects of the etch profile, including sidewall verticality, hole spacing, and structural distortion. 
For confidentiality reasons, the exact variable names and values cannot be disclosed, and thus we present the input and output variables in anonymized form (Variable~1, Variable~2, \ldots, QoI~1, \ldots). 

In this problem, in addition to the available experimental data, field experts provided partial information about the monotonic relationships between the input variables and the quantities of interest (QoIs). 
Table~\ref{tab:tab:mono-redacted} summarizes the redacted version of this expert-provided monotonicity information, where rows correspond to \emph{Variable~1\,--\,17} and columns correspond to \emph{QoI~1\,--\,5}. 
Each entry is marked as increasing ($+1$), decreasing ($-1$), or unknown (0). 
The objective of this problem is to construct accurate surrogate models for all five QoIs under severe data scarcity by integrating the 34 experimental measurements with the monotonicity knowledge provided by field experts.

To test the effectiveness of our proposed RBF-Gen method, we compared the standard RBF method which was trained purely from the dataset with the RBF-Gen method that uses both the dataset and the monotonicity information to train the surrogate model. For both methods, we combined with the standard partial least square method to reduce the input space to 5 dimensions to improve their effectiveness.

In this problem, we employ a Leave-Two-Out (L2O) cross-validation strategy to evaluate the predictive accuracy of the surrogate models. 
For \(n = 34\) experimental samples, there are \(\binom{34}{2} = 561\) folds. 
In each fold, the model is trained on \(n - 2 = 32\) samples and evaluated on the two held-out samples, yielding a total of \(2 \times 561 = 1122\) predictions per quantity of interest (QoI). 
We report the \textit{Average Relative Error} (ARE) and \textit{Average Absolute Error} (AAE), computed by aggregating the \(1122\) predictions across all folds. 
Table~\ref{tab:l2o-results} summarizes the L2O errors for the two surrogate modeling approaches.

The results show that with only \(34\) experiments, purely data-driven surrogates exhibit limited predictive accuracy, while incorporating domain knowledge through the proposed RBF--Gen framework guides the surrogate toward physically consistent trends.
Under the L2O evaluation, clear accuracy improvements are observed for three QoIs (with reductions in both ARE and AAE), while performance on the remaining QoIs remains comparable to the baseline. 
These results demonstrate that expert-informed priors can provide an effective means of enhancing surrogate reliability in data-scarce settings.

\section{Conclusion}
\label{sec: conclusion}

In this work, we introduced \textbf{RBF-Gen}, a knowledge-guided surrogate modeling framework that integrates sparse data with engineering priors. By leveraging the null space of an overparameterized RBF system and training a generator network inspired by InfoGAN, RBF-Gen enables the generation of function ensembles that simultaneously interpolate scarce training data and conform to expert-specified physical properties. Numerical experiments on cantilever beam and shell optimization problems, as well as a real-world semiconductor manufacturing case, demonstrated that RBF-Gen can significantly improve predictive accuracy and design optimization performance compared to standard RBF surrogates, particularly in data-scarce regimes.

A key limitation of the current framework lies in its sensitivity to parameter settings. The effectiveness of RBF--Gen depends on careful tuning of hyperparameters such as the number of RBF centers, null-space scaling, and penalty weights, which may limit its usability in practice. Another limitation is the absence of a mechanism to incorporate other forms of domain knowledge, such as periodic or oscillatory priors commonly encountered in mechanical and control systems. In addition, the accuracy of the proposed method depends on the quality of the domain knowledge provided, and its sensitivity to inaccuracies or inconsistencies in this knowledge has not yet been investigated.

For future work, it is essential to develop more principled heuristics or adaptive strategies for selecting and controlling these parameters to reduce reliance on manual tuning. In addition, extending the framework to encode a broader range of functional priors, including oscillatory and symmetric behaviors, represents an important step toward enhancing the generality and applicability of RBF--Gen. 
Another important direction is extending RBF-Gen beyond surrogate construction to directly enable uncertainty prediction of optimal design's performance. In addition, an interesting and promising direction for future research is to explore the incorporation of physics-informed or variational discriminators, thereby extending the proposed RBF--Gen framework into a generative adversarial formulation to further enhance the flexibility and expressiveness of the model.

Overall, this study highlights the potential of combining scarce data with engineering knowledge to construct reliable surrogates for complex mechanical and manufacturing process design problems, and points the way toward more automated and robust knowledge-guided surrogate modeling frameworks.


\section*{Funding Data}

\begin{itemize}
\item This work was supported by Samsung Electronics Co., Ltd. (Fund number: IO241018-11024-01)
\end{itemize}

\bibliography{sample}
\end{document}